# Towards Terrain-Based Navigation Using Side-Scan Sonar

Ellen Davenport, Junsu Jang, and Florian Meyer

Scripps Institution of Oceanography and Department of Electrical and Computer Engineering
University of California San Diego, La Jolla, CA
Email: {edavenport, jujang, flmeyer}@ucsd.edu

*Abstract*—This paper introduces a statistical model and corresponding sequential Bayesian estimation method for terrain-based navigation using sidescan sonar (SSS) data. The presented approach relies on slant range measurements extracted from the received ping of a SSS. In particular, incorporating slant range measurements to landmarks for navigation constrains the location and altitude error of an autonomous platform in GPS-denied environments. The proposed navigation filter consists of a prediction step based on the unscented transform and an update step that relies on particle filtering. The SSS measurement model aims to capture the highly nonlinear nature of SSS data while maintaining reasonable computational requirements in the particle-based update step. For our numerical results, we assume a scenario with a surface vehicle that performs SSS and compass measurements. The simulated scenario is consistent with our current hardware platform. We also discuss how the proposed method can be extended to autonomous underwater vehicles (AUVs) in a straightforward way and why the combination of SSS sensor and compass is particularly suitable for small autonomous platforms.

*Index Terms*—Side-scan sonar, Bayesian estimation, autonomous vehicles, navigation, marine robotics, particle filtering.

## I. INTRODUCTION

Autonomous platforms have the potential to enable inexpensive and safe data collection at sea with high resolution in time and space. New approaches for determining position underwater will form the basis for longer and more complex unmanned navigation tasks in GPS-denied environments. Of particular interest are localization methods that can be deployed on small autonomous underwater vehicles (sAUVs), i.e., methods that are inexpensive, lightweight, and efficient. SSS is a common payload on sAUVs as it is often used for hydro-geologic surveys and mapping of the seabed [1]–[4]. Due to its ability to generate large, high-resolution images of the seabed and its small form factor, SSS is a promising sensor for sAUV navigation. While navigation using SSS has been attempted several times, algorithmic solutions for efficient sAUV navigation remain unavailable.

In scenarios where the global positioning system (GPS) cannot be used, accurate navigation typically relies on dedicated sensors for dead reckoning, such as an inertial measurement unit (IMU) or a Doppler velocity log (DVL), both of which can be cost and size-prohibitive [1], [4]. There also exist a variety of active acoustic ranging techniques such as long baseline (LBL) and short baseline (SBL) [5], [6], which require the deployment of transponders in the form of moorings.

If a map of relevant features at the seabed is available, SSS sensors have the potential to provide accurate position information. However, computing position from SSS data is challenged by the highly nonlinear nature of the SSS measurement model and a substantial variability of sensor performance due to platform motion and background noise.

### A. sAUV Navigation

When navigating autonomously in the ocean, sAUVs often localize themselves by surfacing to receive GPS signals or by triangulating using active acoustic ranging techniques. In the time between these types of absolute position updates, sAUVs perform relative position updates by dead reckoning [1], [7]. Unfortunately, dead reckoning is always subject to position errors that grow over time, leading to time and depth-limited missions. In addition, using active ranging for sAUV navigation requires the deployment of a transponder infrastructure in a small predefined area of interest and consequently limits the overall scope of the mission to that area [1]. To overcome these limitations, there is a need for new sAUV navigation strategies that rely on a combination of small and inexpensive onboard sensors and sophisticated signal processing and parameter estimation methods.

Terrain-based navigation is an emerging research area in underwater robotics that aims to develop approaches that bound position errors without relying on the deployment of transponder infrastructure. Here, identifiable seabed features are used as landmarks for absolute positioning [2]. Terrain-based navigation assumes that the area of interest has been surveyed at least once, i.e., either the sAUV has performed a first pass while on the surface, the mission of the sAUV is supported by a lead vehicle [4], or previously generated maps are available. A significant ongoing research effort on sonar image segmentation and object classification aims at finding landmarks in SSS surveys [8]–[10].

Early approaches to navigation with SSS, such as in [1], [2], rely on mosaicing of SSS measurements and stochastic maps to track the vehicle state and landmark locations. Smoothing, i.e., the combination of a forward and backward filter, is used to refine current and past position estimates by using SSS data. Here, SSS data is accumulated to an image before being used for navigation. This results in an update rate of the navigation method that is much lower than the rate at which SSS measurements are acquired [1], [2]. The method in [3] fuses onboard inertial sensors of an AUV and SSS data to correct current and past position estimates. It assumes the presence of a support vehicle that performs ranging with respect to the AUV and has access to GPS position. In [4], in-situ feature maps are generated using SSS and matched against a detailed previously generated map. The discrepancy between the two maps is used to refine the position estimate of the vehicle. All the methods listed above have a limited update rate due to the use of entire sonar images or a limited range by dependence on transponder infrastructure.

*B. Contributions*

The fundamental question addressed in this paper is the feasibility of developing a navigation filter for sAUVs that relies on SSS data. We aim to develop models capturing the high-dimensional nature of SSS images while remaining simple enough to be used in a resource-efficient navigation filter. Much of the existing research in this area employs scan-matching and batch processing. In another example, [3] implements a fusion of measurements extracted from SSS with the internal sensors of an AUV by extended Kalman filtering. We aim to improve the vehicle's positioning relative to known landmark locations by developing a particle-based navigation filter that more accurately captures the nonlinearity inherent to SSS measurements. Contrary to existing work, each line of pixels in an SSS image (i.e., each received sonar ping) is considered a separate measurement. Compared to state-of-the-art methods for AUV navigation based on SSS data, this processing approach significantly increases the update rate of the navigation filter. In particular, it makes it possible to obtain an updated position estimate at the ping rate of the SSS.

In this paper, we develop a statistical model and corresponding sequential Bayesian estimation method for terrain-based navigation using SSS data. The presented approach relies on slant range measurements extracted from a SSS ping. Incorporating slant range measurements relative to landmarks for navigation constrains an autonomous platform's location and altitude error in GPS-denied environments. The proposed navigation filter consists of a prediction step based on the unscented transform and an update step that relies on particle filtering. The SSS measurement model aims to capture the highly nonlinear nature of SSS data while maintaining reasonable computational requirements in a particle-based update step. We also introduce a model that can generate synthetic sonar images based on a vehicle trajectory and known landmarks.

The key contributions of this paper are summarized as follows.

- We establish a new model for SSS measurements that describes individual returns rather than entire images.
- We develop a sequential Bayesian estimation method for terrain-based navigation using SSS.

The remaining sections are organized as follows. Section II describes the state transition and measurement models. Section III discusses the generation of simulated images and image preprocessing. Section IV presents the proposed navigation filter. Finally, Section V presents numerical results from our simulation that demonstrate localization accuracy and implementation feasibility. The conclusion of the paper can be found in Section VI along with a discussion of future work.

## II. SYSTEM MODEL

This section will establish the statistical model for terrain-based navigation using SSS.

*A. State Transition Model*

At discrete time step $k$, the 4-D state of the vehicle, $\boldsymbol{x}_k$, is defined as $\boldsymbol{x}_k = [x_k \ y_k \ \theta_k \ \gamma_k]^\mathrm{T}$ where $[x_k \ y_k]^\mathrm{T}$ is the 2-D position in a Cartesian coordinate system, $\theta_k$ is the heading in radians, and $\gamma_k$ is altitude above the seafloor. The control input to the platform is denoted as $\boldsymbol{u}_k = [u_{\mathrm{s},k} \ u_{\mathrm{t},k}]^\mathrm{T}$, where $u_{\mathrm{s},k}$ is the speed and $u_{\mathrm{t},k}$ is the turn rate. The transition from the state at time $k-1$ to time $k$, is described by a transition model $\boldsymbol{x}_k = g(\boldsymbol{x}_{k-1}, \boldsymbol{n}_k, \boldsymbol{u}_k)$ that includes the driving noise vector $\boldsymbol{n}_k = [n_{\mathrm{s},k} \ n_{\mathrm{t},k} \ n_{\theta,k} \ n_{\gamma,k}]^\mathrm{T}$. The elements of the driving noise vector $\boldsymbol{n}_k$ are statistically independent and Gaussian distributed with variances $\sigma_\mathrm{s}^2$, $\sigma_\mathrm{t}^2$, $\sigma_\theta^2$, and $\sigma_\gamma^2$. Furthermore, driving noise vectors $\boldsymbol{n}_k$ are statistically independent across time $k$. The functional form of $g(\boldsymbol{x}_{k-1}, \boldsymbol{n}_k, \boldsymbol{u}_k)$ is given by

$$\begin{bmatrix} x_k \\ y_k \\ \theta_k \\ \gamma_k \end{bmatrix} = \begin{bmatrix} x_{k-1} - \frac{v_{\mathrm{s},k}}{v_{\mathrm{t},k}} \sin(\theta_{k-1}) + \frac{v_{\mathrm{s},k}}{v_{\mathrm{t},k}} \sin(\theta_{k-1} + v_{\mathrm{t},k} \Delta_k) \\ y_{k-1} + \frac{v_{\mathrm{s},k}}{v_{\mathrm{t},k}} \cos(\theta_{k-1}) - \frac{v_{\mathrm{s},k}}{v_{\mathrm{t},k}} \cos(\theta_{k-1} + v_{\mathrm{t},k} \Delta_k) \\ \theta_{k-1} + v_{\mathrm{t},k} \Delta_k + n_{\theta,k} \Delta_k \\ \gamma_{k-1} + n_{\gamma,k} \end{bmatrix}$$
(1)

where we introduce the short notation $v_{\mathrm{s},k} = u_{\mathrm{s},k} + n_{\mathrm{s},k}$ and $v_{\mathrm{t},k} = u_{\mathrm{t},k} + n_{\mathrm{t},k}$. Note that $\Delta_k$ is the time duration between time steps $k-1$ and $k$. The model for the first three elements of $\boldsymbol{x}_k$ is developed in [11]. At each step, the control inputs for speed and turn rate that determine the new position and heading are corrupted by additive Gaussian noise. The vehicle altitude is simply the altitude from the previous time step corrupted by additive noise. A potential limitation of the current model is that it does not consider ocean currents. Developing a more sophisticated state transition model incorporating ocean currents is subject to future work.

*B. Measurement Model*

SSS instruments transmit acoustic pulses ("pings") and generate an image from the signal returns that are backscattered by features on the seafloor. Features that are rough backscatter more power compared to features that are soft [12]. Each ping is associated with a new line of pixels in the image. The sonar transducers are mounted or towed such that they provide "cross-track" measurements, i.e., measurements that are perpendicular to the direction of motion. Typically, there are two sonar transducers so that measurements can be performed on the port and starboard side of the platform. Fig. 1 shows the geometry of transducers, water column, and seafloor. Each pixel in a SSS image corresponds to the total intensity reflected within a particular area of the seafloor. The size of the area corresponding to each pixel and the effective range of the SSS depends on the hardware configuration (e.g., transmit frequency) and altitude above the seafloor. The random variability of acoustic propagation, including a change in the sea state or the bottom topography [12], affect the quality of SSS measurements. The sensor generates an image by stitching measurements from each time step $k$ in the direction of motion. Each time step is one line of pixels in this image [12]. Each pixel corresponds to the range from the vehicle to the ocean bottom through the water column. As shown in Fig. 1, this "slant range" is the hypotenuse of a triangle [13].

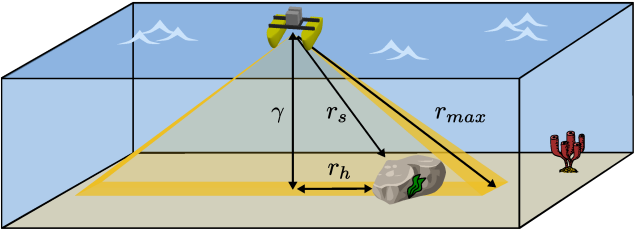

Fig. 1: The SSS takes measurements perpendicular to the direction of motion of the sensing platform. The maximum range of the sonar sensor, $r_{\max}$, defines the hypotenuse of the triangle formed between the vehicle and the farthest detection point of the SSS. When a landmark is present, its respective slant range, $r_s$, can be extracted from the sonar image. This slant range corresponds to a horizontal range, $r_h$. The vehicle altitude above the seafloor, $\gamma$, is needed to obtain $r_h$ from $r_s$.

Fig. 2 shows a SSS image that combines returns from a port and starboard transducer. The dark section at the center of the image is called the *nadir* and is proportional to the time of the first acoustic return, often referred to as the first bottom return. The size of the nadir can be used to measure the altitude above the seafloor [14]. Assuming that a map of the features at the seafloor has been established, features in SSS images can be used as landmarks for terrain-based navigation.

Each landmark with index $d \in \{1, \ldots, D\}$ is represented by a rectangle described by the vector $\boldsymbol{m}_d = [x_d \ y_d \ \theta_d \ l_d \ w_d]^\mathrm{T}$. Here, $[x_d \ y_d]^\mathrm{T}$ denotes the 2-D position of the center of the landmark, and $\theta_d$ is the landmark orientation. $\theta_d$ is in radians and is 0 when the vehicle points directly east and $\frac{\pi}{2}$ when the vehicle points north. In addition, $[l_d \ w_d]^\mathrm{T}$ denotes the length and width of the landmark.

As further discussed in Section III, each landmark can generate two distances measurements in the slant range, $\boldsymbol{z}_{k,d} = \begin{bmatrix} z_{k,d}^{(1)} & z_{k,d}^{(2)} \end{bmatrix}^\mathrm{T}$. When landmark $d \in \{1, \ldots, D\}$ is detected by the SSS, these distances are extracted from the image and comprise the landmark measurement. If landmark with index $d \in \{1, \ldots, D\}$ is not detected at time step $k$, the corresponding measurement is set to $\boldsymbol{z}_{k,d} = \boldsymbol{z}_{\max}$. Here, we have introduced $\boldsymbol{z}_{\max} = \begin{bmatrix} r_{\max} & r_{\max} \end{bmatrix}^\mathrm{T}$. Using extracted slant ranges instead of the entire set of pixels as measurements significantly reduces the computational complexity of the navigation filter discussed in Section V.

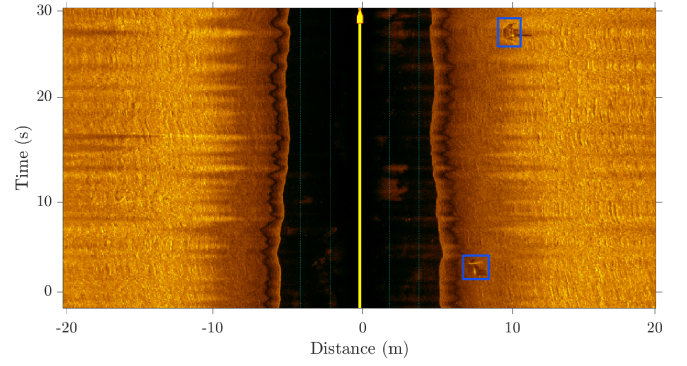

Fig. 2: SSS image with two landmarks. The horizontal axis is a slant range perpendicular to the vehicle track. The negative distance represents the port side, and the positive distance represents the starboard side of the vehicle. This image shows approximately 30 seconds of data, and the maximum slant range at each side is 20 m. The center frequency of the transmitted pulse is 1800 kHz, and the ping rate is 20 Hz, meaning that each line of pixels is 50 ms apart. The illumination of each pixel represents the intensity of the acoustic signal reflected by the seafloor. This data was collected at the Scripps Institute of Oceanography using the sensing platform and artificial landmark shown in Fig. 3.

At time $k$, the SSS measurement covers a line segment on the seafloor that has port and starboard end points defined by

$$\boldsymbol{x}_k^\mathrm{p} = \begin{bmatrix} x_k \\ y_k \end{bmatrix} + \begin{bmatrix} \cos(\theta_k + \frac{\pi}{2}) \\ \sin(\theta_k + \frac{\pi}{2}) \end{bmatrix} \sqrt{r_{\max}^2 - \gamma_k^2}$$

$$\boldsymbol{x}_k^\mathrm{s} = \begin{bmatrix} x_k \\ y_k \end{bmatrix} - \begin{bmatrix} \cos(\theta_k + \frac{\pi}{2}) \\ \sin(\theta_k + \frac{\pi}{2}) \end{bmatrix} \sqrt{r_{\max}^2 - \gamma_k^2} \quad (2)$$

where $r_{\max}$ is the maximum slant range of the SSS. Line segments $s \in \{1, 2, 3, 4\}$ represent the sides of a rectangular landmark $d \in \{1, \ldots, D\}$ on the seafloor. The sides $s \in \{1, 2, 3, 4\}$ correspond to the landmark's south, west, north, and east sides. They are defined by the beginning and end points

$$\boldsymbol{x}_{d,s}^\mathrm{b} = \begin{bmatrix} x_d \\ y_d \end{bmatrix} + b_s^{(1)} \frac{1}{2} \begin{bmatrix} l_d(\sin(\theta_d) - \cos(\theta_d)) \\ -w_d(\sin(\theta_d) + \cos(\theta_d)) \end{bmatrix}$$

$$\boldsymbol{x}_{d,s}^\mathrm{e} = \begin{bmatrix} x_d \\ y_d \end{bmatrix} + b_s^{(2)} \frac{1}{2} \begin{bmatrix} l_d(\sin(\theta_d) + \cos(\theta_d)) \\ w_d(\sin(\theta_d) - \cos(\theta_d)) \end{bmatrix} \quad (3)$$

with $\boldsymbol{b}_s = \begin{bmatrix} b_s^{(1)} & b_s^{(2)} \end{bmatrix}^\mathrm{T}$, $s \in \{1, 2, 3, 4\}$ given by $\boldsymbol{b}_1 = [1 \ 1]^\mathrm{T}$, $\boldsymbol{b}_2 = [1 \ -1]^\mathrm{T}$, $\boldsymbol{b}_3 = [-1 \ -1]^\mathrm{T}$, and $\boldsymbol{b}_4 = [-1 \ 1]^\mathrm{T}$.

Let $h_1(\boldsymbol{m}_d, \boldsymbol{x}_k)$ be the binary function that is equal to 1 if the line segments defined in (2) intersect with the line segments defined in (3), and zero otherwise. Whenever there is any intersection, there will always be two intersection points. For the case $h_1(\boldsymbol{m}_d, \boldsymbol{x}_k) = 1$, let $h_2(\boldsymbol{m}_d, \boldsymbol{x}_k)$ be the function that provides the resulting two intersection points on the seafloor, $\begin{bmatrix} \boldsymbol{x}_{k,d}^{(1)} & \boldsymbol{x}_{k,d}^{(2)} \end{bmatrix}^\mathrm{T}$. The measurement model for 2-D range and width measurements of landmark $d$ can now be written as $\boldsymbol{z}_{k,d} = h(\boldsymbol{x}_k, \boldsymbol{m}_d, \boldsymbol{n}_{k,d})$, where $\boldsymbol{n}_{k,d}$ is zero-mean

Gaussian measurements noise with covariance matrix $\sigma^2 \boldsymbol{I}_2$, and $h(\boldsymbol{x}_k, \boldsymbol{m}_d, \boldsymbol{n}_{k,d})$ is given by

$$h(\boldsymbol{x}_k, \boldsymbol{m}_d, \boldsymbol{n}_{k,d}) = \begin{cases} h_3(h_2(\boldsymbol{m}_d, \boldsymbol{x}_k), \boldsymbol{x}_k) + \boldsymbol{n}_{k,d} & h_1(\boldsymbol{m}_d, \boldsymbol{x}_k) = 1 \\ \boldsymbol{z}_{\max} & h_1(\boldsymbol{m}_d, \boldsymbol{x}_k) = 0. \end{cases} \quad (4)$$

Here, we have introduced the function $[r_{k,d}^{(1)}\ r_{k,d}^{(2)}]^{\mathrm{T}} = h_3(\boldsymbol{x}_{k,d}^{(1)}, \boldsymbol{x}_{k,d}^{(2)}, \boldsymbol{x}_k)$, that provides the distances of the two intersection points with respect to vehicle state $\boldsymbol{x}_k$ in slant range, i.e.,

$$\begin{bmatrix} r_{k,d}^{(1)} \\ r_{k,d}^{(2)} \end{bmatrix} = \begin{bmatrix} \sqrt{\|\boldsymbol{x}_{k,d}^{(1)} - [x_k\ y_k]^{\mathrm{T}}\|^2 + \gamma_k^2} \\ \sqrt{\|\boldsymbol{x}_{k,d}^{(2)} - [x_k\ y_k]^{\mathrm{T}}\|^2 + \gamma_k^2} \end{bmatrix}. \quad (5)$$

The measurement noise $\boldsymbol{n}_{k,d}$ is assumed statistically independent across $k$ and $d$. In summary, $h_1(\boldsymbol{m}_d, \boldsymbol{x}_k)$ indicates the presence of a landmark, $h_2(\boldsymbol{m}_d, \boldsymbol{x}_k)$ provides the points of intersection with said landmark, and $h_3(\boldsymbol{x}_{k,d}^{(1)}, \boldsymbol{x}_{k,d}^{(2)}, \boldsymbol{x}_k)$ provides the slant range to the landmark based on the output from $h_2$. The likelihood function corresponding to the measurement model in (5) is given by

$$p(\boldsymbol{z}_{k,d}|\boldsymbol{x}_k) = \begin{cases} \mathcal{N}(h_3(h_2(\boldsymbol{m}_d, \boldsymbol{x}_k), \boldsymbol{x}_k); \sigma^2 \boldsymbol{I}_2) & h_1(\boldsymbol{m}_d, \boldsymbol{x}_k) = 1 \\ \delta(\boldsymbol{z}_{k,d} - \boldsymbol{z}_{\max}) & h_1(\boldsymbol{m}_d, \boldsymbol{x}_k) = 0 \end{cases}. \quad (6)$$

Here, $\delta(\cdot)$ is the unit pulse that is one for $\delta(\boldsymbol{z}_{k,d} - \boldsymbol{z}_{\max})$ equal to zero, and zero otherwise. Note that in (6), the case $h_1(\boldsymbol{m}_d, \boldsymbol{x}_k) = 0$ can provide information on the vehicle state if the true measurement does indicate that a landmark is present. In other words, the presence and absence of a landmark are both informative.

In addition to the $D$ measurements performed with respect to landmarks, the altitude above the seafloor and the heading are measured as $z_{k,\mathrm{a}} = \gamma_k + n_{k,\mathrm{a}}$ and $z_{k,\mathrm{c}} = \theta_k + n_{k,\mathrm{c}}$. Here, $n_{k,\mathrm{a}}$ and $n_{k,\mathrm{c}}$ are zero-mean measurements noises with variances $\sigma_\mathrm{a}^2$ and $\sigma_\mathrm{c}^2$. The measurement noises $n_{k,\mathrm{a}}$ and $n_{k,\mathrm{c}}$ are statistically independent with respect to each other and across time $k$. They are also statistically independent of measurement noises $\boldsymbol{n}_{k,d}$ for all $k$ and $d$. The joint measurement of length $2D + 2$ at time $k$, is given by $\boldsymbol{z}_k = [\boldsymbol{z}_{k,1} \ldots \boldsymbol{z}_{k,D}\ z_{k,\mathrm{a}}\ z_{k,\mathrm{c}}]^{\mathrm{T}}$. The corresponding joint likelihood function factorizes as

$$p(\boldsymbol{z}_k|\boldsymbol{x}_k) = p(z_{k,\mathrm{c}}|\boldsymbol{x}_k) p(z_{k,\mathrm{a}}|\boldsymbol{x}_k) \prod_{d=1}^{D} p(\boldsymbol{z}_{k,d}|\boldsymbol{x}_k). \quad (7)$$

This likelihood function will be used in the navigation filter developed in Section IV.

### III. SYNTHETIC SONAR DATA AND PROCESSING

We developed a forward model for simulation that generates sonar data given a vehicle track, sonar parameters, and landmark locations. For each vehicle position on the track, we create a line of pixels representing one received sonar ping. Pixel values are binary, i.e., pixels with no landmark are 0's while pixels containing a landmark are 1's. Similarly to related work [10], we assume a relatively flat sea bottom. While the flight height can change, the cross-track profile is assumed flat (i.e., the depth on the left and right sides of the vehicle are the same) at all times. Lines of pixels can be concatenated into a synthetic sonar image for visualization purposes. However, each line is used individually to extract measurements for the proposed navigation filter. The synthetically generated sonar data represents a real sonar image preprocessed by an image classifier that detects relevant landmarks. As further discussed in Section II, potential errors in the preprocessing stage are modeled by additive noise.

For each line of pixels of a synthetic or real sonar image, a slant range and the length measurement is extracted for each landmark that is present. Measurements are extracted by traversing the line of pixels until an edge is detected. This edge is marked as the near side of a landmark. The landmark's far side is further obtained by detecting the following edge. Since we get range and length measurements directly from the pixel indexes related to edges, the obtained measurements are in the slant range. This process generates between 0 and $D$ SSS measurements, depending on how many landmarks are present in the line of pixels. The remaining landmark measurements are set to $\boldsymbol{z}_{\max}$ to indicate that the corresponding landmark was not present in the line of pixels. Measurements of landmarks that are not present in the line of pixels are set to $\boldsymbol{z}_{\max}$. The vector that consists of the resulting $D$ landmark measurements, $\boldsymbol{z}_k$, is the result of this processing stage and will be used at each update step of the navigation filter as discussed in IV-B. Future work will be focused on processing real data collected from a surface vehicle, which was developed in collaboration with *Seafloor Systems, Inc*. A picture of this platform taken at the *Scripps Institution of Oceanography* is shown in Fig. 3.

### IV. THE NAVIGATION FILTER

At time $k$, we aim to estimate the state, $\boldsymbol{x}_k$, of the AUV from all measurements $\boldsymbol{z}_{1:k}$. Given the conditional probability density function (PDF) of the state, $p(\boldsymbol{x}_k|\boldsymbol{z}_{1:k})$, the minimum mean-squared error (MMSE) estimate [15] of the state can be obtained as

$$\hat{\boldsymbol{x}}_k^{\mathrm{MMSE}} = \int \boldsymbol{x}_k p(\boldsymbol{x}_k|\boldsymbol{z}_{1:k}) \mathrm{d}\boldsymbol{x}_k. \quad (8)$$

A Bayes filter [16], which consists of a prediction and update steps, is applied to compute an approximation of the conditional PDF $p(\boldsymbol{x}_k|\boldsymbol{z}_{1:k})$. The prediction step uses the Chapman-Kolmorogov equation, which involves the state-transition function in (1). The update step is based on Baye's rule and the likelihood function in (7). Due to the nonlinearities in our state-transition and measurement models, we use sigma points [17], [18] in the prediction step and random samples "particles" [16] in the update step for the computation of an approximate mean $\boldsymbol{\mu}_k$ and covariance matrix $\boldsymbol{C}_k$ of $p(\boldsymbol{x}_k|\boldsymbol{z}_{1:k})$. Using particle-based instead of sigma points

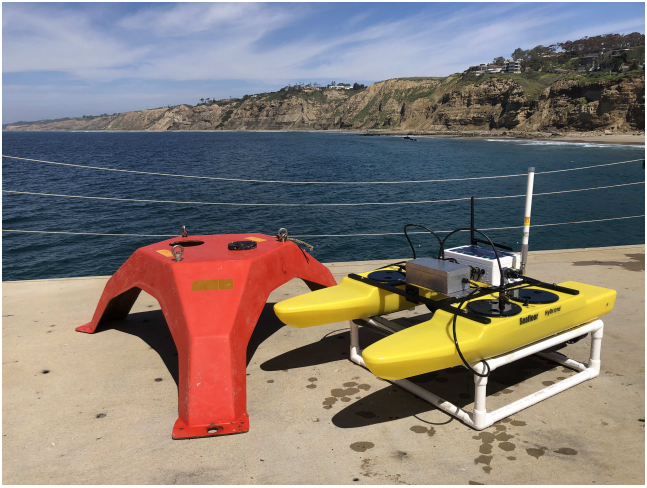

Fig. 3: *Hydrone* surface vehicle with integrated *Ark Scout Mk II* SSS transducers shown next to an artificial landmark. Two of these landmarks were used to generate the data in Fig. 2. This picture was taken at the Ellen Browning Scripps Memorial Pier in La Jolla, CA. Data collection was conducted close to the pier.

$$\boldsymbol{x}_{k-1}^{i+N} = \boldsymbol{\eta}_{k-1} - (\sqrt{N\boldsymbol{\Sigma}_{k-1}})^i \quad i \in \{N+1,\ldots,2N\}$$

The weights corresponding to these sigma points are set as $w^i = \frac{1}{2N}$, $i \in \{1,\ldots,2N\}$.

Next, 8-D sigma points are passed through the state transition model in (1), i.e.,

$$\underline{\boldsymbol{x}}_k^{i-} = g(\underline{\boldsymbol{x}}_{k-1}^i, \overline{\boldsymbol{x}}_{k-1}^i, \boldsymbol{u}_k) \quad i \in \{1,\ldots,2N\}.$$

Here, $\underline{\boldsymbol{x}}_{k-1}^i$ and $\overline{\boldsymbol{x}}_{k-1}^i$ denote the first four and last four elements of sigma point $\boldsymbol{x}_k^i$. Finally, an updated 4-D mean, and covariance are computed according to

$$\boldsymbol{\mu}_k^- = \sum_{i=1}^{2N} w^i \underline{\boldsymbol{x}}_k^{i-}$$

$$\boldsymbol{C}_k^- = \sum_{i=1}^{2N} w^i (\underline{\boldsymbol{x}}_k^{i-} - \boldsymbol{\mu}_k^-)(\underline{\boldsymbol{x}}_k^{i-} - \boldsymbol{\mu}_k^-)^{\mathrm{T}}.$$

This mean $\boldsymbol{\mu}_k^-$ and covariance matrix $\boldsymbol{C}_k^-$ represent the predicted posterior PDF $p(\boldsymbol{x}_k|\boldsymbol{z}_{1:k-1}) \approx \mathcal{N}(\boldsymbol{x}_k; \boldsymbol{\mu}_k^-, \boldsymbol{C}_k^-)$.

requires more computational resources but makes it possible to obtain accurate results even if the underlying model is strongly nonlinear. In the considered problem, the nonlinearity in the state-transition model is moderate. The more efficient computation based on sigma points can thus also provide accurate results and is preferred. Due to (8), the approximate mean computed by the filter, $\boldsymbol{\mu}_k$, is at the same time an approximation of the MMSE estimate, i.e., $\hat{\boldsymbol{x}}_k^{\mathrm{MMSE}} \approx \boldsymbol{\mu}_k$. The flow chart in Fig. 4 shows the overall filter architecture. The prediction and update steps are described next.

### A. Prediction Step

We use sigma points in the prediction step of the navigation filter because the state transition model is nonlinear. Let, $\mathcal{N}(\boldsymbol{x}_{k-1}^i; \boldsymbol{\mu}_{k-1}, \boldsymbol{C}_{k-1})$ be a Gaussian approximation of the marginal posterior PDF $p(\boldsymbol{x}_{k-1}^i|\boldsymbol{z}_{1:k-1})$ that has been computed at the previous time step $k-1$. First, we introduce the 8-D augmented mean $\boldsymbol{\eta}_{k-1} = [\boldsymbol{\mu}_{k-1}^{\mathrm{T}}\ 0\ 0\ 0\ 0]^{\mathrm{T}}$ and the corresponding covariance matrix

$$\boldsymbol{\Sigma}_{k-1} = \begin{bmatrix} \boldsymbol{C}_{k-1} & \boldsymbol{0} \\ \boldsymbol{0} & \mathrm{diag}(\boldsymbol{\sigma}^2) \end{bmatrix} \quad (9)$$

where $\mathrm{diag}(\boldsymbol{\sigma}^2)$ denotes the diagonal matrix with diagonal elements given by the vector $\boldsymbol{\sigma}^2 = [\sigma_s^2\ \sigma_t^2\ \sigma_\theta^2\ \sigma_\gamma^2]^{\mathrm{T}}$. Note that this mean vector and covariance matrix have been augmented by the means and variances of $\boldsymbol{n}_k$. This augmentation makes using sigma points in the prediction step possible despite the nonlinear relationship between $\boldsymbol{n}_k$ and $\boldsymbol{x}_{k-1}$. The length of the new state vector is $N=8$. Sigma points are computed from the mean state according to [17] and [18]. In effect, $2N$ sigma points are evenly spaced around one sigma point at the current augmented state estimate. The augmented sigma points, $\boldsymbol{x}_{k-1}^i$, are defined as

$$\boldsymbol{x}_{k-1}^i = \boldsymbol{\eta}_{k-1} + (\sqrt{N\boldsymbol{\Sigma}_{k-1}})^i \quad i \in \{1,\ldots,N\}$$

### B. Update Step

We use a particle filter to compute the updated posterior PDF $p(\boldsymbol{x}_k|\boldsymbol{z}_{1:k}) \approx \mathcal{N}(\boldsymbol{x}_k; \boldsymbol{\mu}_k, \boldsymbol{C}_k)$ that takes all available measurements into account. Due to the highly nonlinear measurement model defined in (4), a sigma point-based computation is unsuitable for the update step. Importance sampling is performed by directly using the Gaussian representation $\mathcal{N}(\boldsymbol{x}_k; \boldsymbol{\mu}_k^-, \boldsymbol{C}_k^-)$ of the predicted posterior PDF computed in the Section IV-A as a proposal PDF [16], [19]. In other words, we sample $I$ particles denoted as $\{\bar{\boldsymbol{x}}_k^i\}_{i=1}^I$ from $\mathcal{N}(\boldsymbol{x}_k; \boldsymbol{\mu}_k^-, \boldsymbol{C}_k^-)$. Corresponding particle weights are obtained by evaluating the joint likelihood function provided in (7) for each particle, i.e.,

$$w_k^i = p(\boldsymbol{z}_{k,\mathrm{c}}|\boldsymbol{x}_k^i) p(\boldsymbol{z}_{k,\mathrm{a}}|\boldsymbol{x}_k^i) \prod_{d=1}^{D} p(\boldsymbol{z}_{k,d}|\boldsymbol{x}_k^i). \quad (10)$$

Note that the evaluation of (10) can also provide position information if no landmark is detected, i.e., if $\boldsymbol{z}_{k,d} = \boldsymbol{z}_{\max}$ for all $d \in \{1,\ldots,D\}$. According to the second line in (6), for a particle corresponding to a detected landmark, the computed weight according to (10) will be zero, and the particle will be discarded. For numerical stability, the evaluation of the individual factors in (10) is performed in the log domain. Consequently, in the log domain, the product of factors in (10) is computed as a sum. After a normalization step, weights are converted to the linear domain and normalized to one. Finally, a new state estimate and covariance matrix are obtained as [16], [19]

$$\boldsymbol{\mu}_k = \sum_{i=1}^{I} \bar{w}_k^i \boldsymbol{x}_k^i$$

$$\boldsymbol{C}_k = \sum_{i=1}^{I} \bar{w}^i (\boldsymbol{x}_k^i - \bar{\boldsymbol{x}}_k)(\boldsymbol{x}_k^i - \bar{\boldsymbol{x}}_k)^{\mathrm{T}}$$

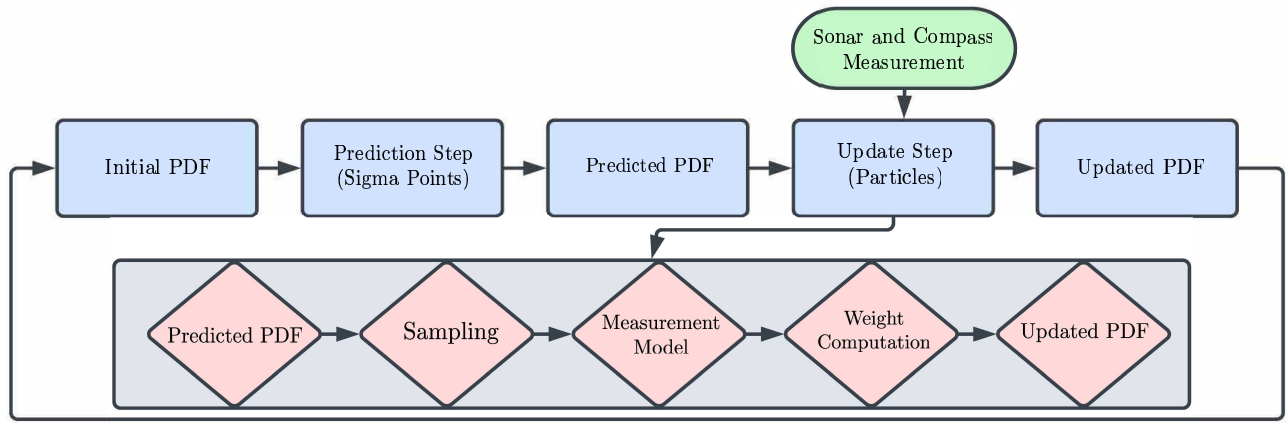

Fig. 4: Flow diagram describing the processes of the proposed navigation filter. Blocks shown in blue depict the main processing steps and their results. The sub-process in red shows details of the proposed particle-based update step.

where $\{\bar{w}_k^i\}_{i=1}^I$ are the weights that have been normalized, i.e., $\bar{w}_k^i = w_k^i / \sum_{i'=1}^I w_k^{i'}$. This mean and covariance matrix define a Gaussian approximation of the updated marginal posterior PDF, $p(\boldsymbol{x}_k^i|\boldsymbol{z}_{1:k}) = \mathcal{N}(\boldsymbol{x}_k^i; \boldsymbol{\mu}_k, \boldsymbol{C}_k)$. This PDF is then used in the prediction step at time $k+1$. To avoid particle degeneracy, a resampling step has to be performed at certain time steps [16], [19]. The mean, $\boldsymbol{\mu}_k$, is at the same time an approximation of the MMSE state estimate $\hat{\boldsymbol{x}}_k^{\text{MMSE}}$ in (8).

## V. NUMERICAL RESULTS

We evaluate algorithm performance in a simulated scenario with different landmark spacings. As a reference method, we use a navigation filter that does not use SSS measurements. In the state transition model, we set the driving noise variances as $\sigma_s^2 = 1.5\,\text{m}^2/\text{s}^2$, $\sigma_t^2 = 0.5\,\text{rad}^2/\text{s}^2$, $\sigma_\theta^2 = 0.2\,\text{rad}^2$, and $\sigma_\gamma^2 = 0.1\,\text{m}^2$. The measurement model is characterized by $\sigma_r^2 = 2.5\,\text{m}^2$, $\sigma_a^2 = 0.5\,\text{m}^2$, $\sigma_c^2 = 0.2\,\text{rad}^2$. At time $k = 0$, we set $\boldsymbol{\mu}_0$ equal to the true vehicle position and $\boldsymbol{C}_0 = \text{diag}([\sigma_r^2\ \sigma_r^2\ \sigma_a^2\ \sigma_c^2]^\text{T})$. The time duration between time steps is $\Delta_k = 0.1s$. Each simulation run consists of 12000 time steps corresponding to a total duration of 20 minutes.

In the considered scenario, landmarks are located on an evenly-spaced grid. An example scenario is shown in Fig. 5(a). As a performance metric, we use the root mean-square error (RMSE) of the 3-dimensional vehicle location. Fig. 5(a) shows true and estimated vehicle tracks for a single simulation run. A landmark spacing of 50 meters was used. It can be seen that the RMSE related to the position estimates provided by the reference methods keeps growing as time evolves. This is because the reference method relies solely on integrating heading measurements over time and has, thus, no access to absolute position information. At the end of the simulated scenario, the RMSE related to the reference methods is typically more than 10 meters. On the other hand, when landmark spacing is reasonable, the RMSE of the position estimate provided by the proposed method remains bounded. Fig. 5(b) shows a close-up of the very end of the track from Fig. 5(a). It can be seen that the estimation error of the reference method is significant, while the estimated error of the proposed method is quite small.

Fig. 6 shows the RMSE related to the reference method and the proposed navigation filter averaged over 300 simulation runs. Scenarios with 50-meter and 100-meter landmark spacing are considered for the proposed navigation filter. For the reference method, which can not use position information of the landmarks, the average RMSE of a 20-minute trajectory is 8.12 meters. For the proposed navigation filter, the average RMSE decreases to 2.99 meters and 1.56 meters for 100-meter and 50-meter landmark spacing, respectively. Fig. 6 also confirms that when no landmarks are used, the RMSE continues to grow with time while SSS measurements of landmarks can bound the position error. As expected, increased spacing between landmarks results in a more significant estimation error. In this simulation, the probability of the vehicle measuring a landmark at any given time step, $k$, was 1.4% for a landmark spacing of 100 meters. On the other hand, in the scenario with 50 meters spacing, the probability of the vehicle measuring a landmark was 5.7%. As can also be seen in Fig. 6, a 4.3% increase in the probability of measuring a landmark led to a significantly reduced localization error.

## VI. CONCLUSION AND FUTURE WORK

We developed a method that has the potential to increase navigation accuracy for sAUVs with SSS. In particular, the error of position estimates can be reduced by using detections of landmarks in sonar pings as measurements. Our simulation results showed an improved navigation performance compared to a reference method that relies on dead reckoning. In particular, the position error related to the proposed navigation strategy can be bounded if landmarks detections become available regularly. For example, in the scenario with a regular grid of landmarks spaced 50 m appart, the probability of seeing a landmark at any given time was 5.7%. In the scenario with

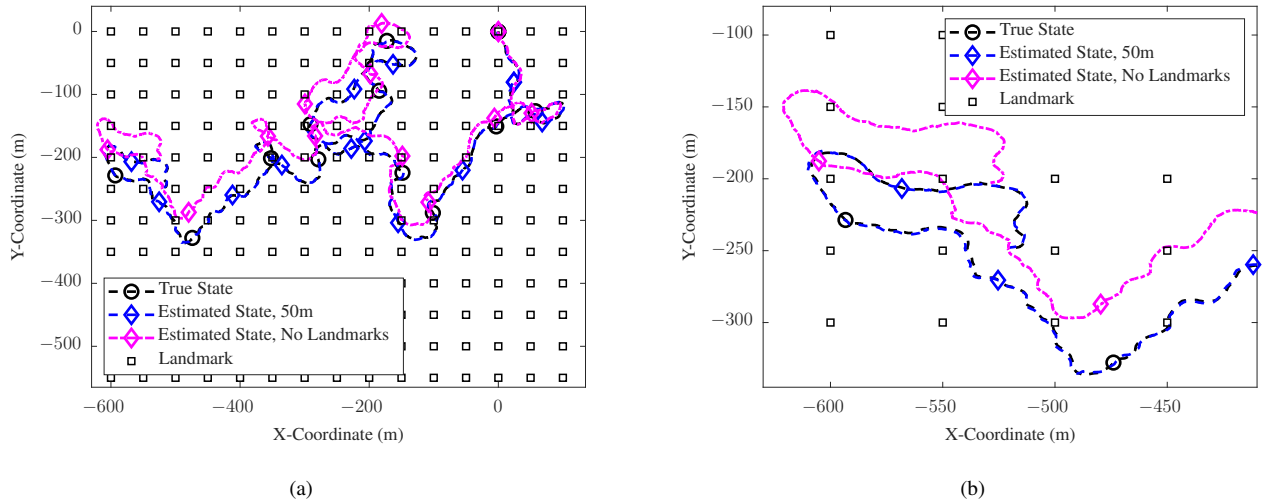

Fig. 5: True and estimated vehicle tracks resulting from a single simulation run. The proposed navigation filter and a reference method that does not use landmark measurements are considered. Landmarks, spaced evenly at 50-meter intervals in the x and y directions, are also shown. The duration of the simulation is 20 minutes. The entire tracks are shown in Fig. 5(a) while a zoom-in is provided in Fig. 5(b). The location error associated with the estimated track obtained by the proposed navigation filter is approximately the same at the end of the track as at the beginning. The location error obtained by the reference method does increase over time.

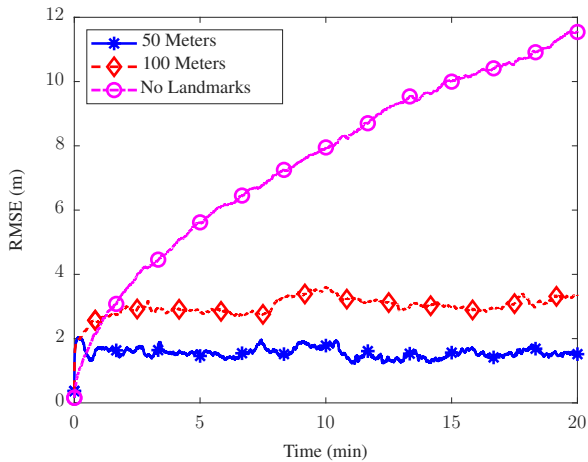

Fig. 6: Average RMSE versus time for position estimates provided by the proposed navigation filter. Two scenarios with different landmark spacings are considered. The RMSE of the reference method that does not use landmark information is also shown. RMSE curves have been averaged over 300 simulation runs.

100 m landmark spacing, this probability dropped to 1.4%. In practice, a higher probability can be obtained by identifying small features as landmarks or deploying artificial landmarks. Finally, the method is computationally efficient, making it possible to use in real-time. The computational load can be adjusted by processing received sonar pings less frequently or using fewer particles. For example, if the state estimate variance is small and the vehicle location is far from known landmarks, received sonar pings can be processed less often.

Results based on real data will allow us to assess the assumed state transition model's accuracy and help quantify the positioning error in realistic scenarios. Real data processing will also provide insights into the frequency of landmark detections necessary to keep the error of position estimates below a desired threshold. Another important consideration is the standard deviation of the range and length measurements, which in practice will depend on the underlying landmark detection algorithm. It will also be essential to determine a robust estimate for these standard deviations for real-world application of the proposed navigation filter. The next steps of this research consist of (1) collecting a large-scale dataset at sea, (2) developing a more realistic state-transition model that also takes ocean current into account, (3) designing a robust proposal distribution for the case where the interval between landmarks is long, and the location information of the vehicle's position has become uninformative. The third task is important in scenarios where a landmark is detected in the current received sonar ping, but the posterior distribution of the vehicle state is so uninformative that no generated particle corresponds to a vehicle position with landmark detection. Instead of using the predicted posterior as the proposal distribution in the update step, an alternative proposal is developed based on current measurements. Promising directions for future research include extending a probabilistic association of slant ranges to landmarks [20]–[22] and embedding deep neural networks for landmark detection [23].

## ACKNOWLEDGEMENT

This work was supported in part by the Office of Naval Research under Grant N00014-21-1-2267 and by the National Science Foundation (NSF) under CAREER Award No. 2146261. The authors thank Mr. Sean Fish and Mr. Matthew Morozov for platform development and testing.